\documentclass[11pt]{article}
\usepackage{mtsummit2019}
\usepackage{times}
\usepackage{url}
\usepackage{latexsym}
\usepackage[small,bf]{caption} 
\usepackage{subcaption}

\usepackage{natbib}

\usepackage{amsmath}
\usepackage{booktabs}
\usepackage{multirow}

\usepackage[vlined,linesnumbered]{algorithm2e}
\usepackage{hyperref}
\hypersetup{colorlinks,citecolor=black, linkcolor=black, urlcolor=black}
	
\usepackage[nameinlink,capitalize]{cleveref}

\newcommand{\x}{\mathbf{x}}
\newcommand{\y}{\mathbf{y}}

\newcommand{\VTheta}{\mathbf{\Theta}}

\DeclareMathOperator*{\argmax}{arg\,max~}

\newcommand{\p}{\mathrm{p}}

\title{Incremental Adaptation of NMT for \\ Professional Post-editors: A User Study}

\author{Miguel Domingo$^1$ \and Mercedes Garc\'ia-Mart\'inez$^2$ \and \'Alvaro Peris$^1$ \and Alexandre Helle$^2$\protect\\ \and \textbf{Amando Estela}$^2$ \and \textbf{Laurent Bi\'e}$^2$ \and \textbf{Francisco Casacuberta}$^1$ \and 
\textbf{Manuel Herranz}$^2$ \\ $^1$PRHLT Research Center - Universitat Polit{\`e}cnica de Val{\`e}ncia \protect\\ \{midobal, lvapeab, fcn\}@prhlt.upv.es \protect\\ $^2$Pangeanic / B.I Europa - PangeaMT Technologies Division\protect\\ \{m.garcia, a.helle, a.estela, l.bie, m.herranz\}@pangeanic.com}

\date{}

\begin{document}
\maketitle
\begin{abstract}
	A common use of machine translation in the industry is providing initial translation hypotheses, which are later supervised and post-edited by a human expert. During this revision process, new bilingual data are continuously generated. Machine translation systems can benefit from these new data, incrementally updating the underlying models under an online learning paradigm. 
	We conducted a user study on this scenario, for a neural machine translation system. The experimentation was carried out by professional translators, with a vast experience in machine translation post-editing. The results showed a reduction in the required amount of human effort needed when post-editing the outputs of the system, improvements in the translation quality and a positive perception of the adaptive system by the users.
\end{abstract}

\section{Introduction}

Translation post-editing is a common use case of machine translation (MT) in the industrial environment. Post-editing consists of the supervision by a human agent of outputs generated by an MT system, who corrects the errors made by the MT system. As MT systems are continuously improving their capabilities, translation post-editing has acquired major relevance in the translation market  \citep{Arenas08,Hu16}. As a byproduct of this process, new data are continuously generated. These data have valuable properties: they are domain-specific training samples, which can be leveraged for adapting the system towards a given domain or post-editor. Moreover, an adaptive system can learn from its mistakes. In other words, it can avoid making the same errors again. 

A typical way of profiting from these post-edits consists in updating the system following an online learning paradigm: as the user validates a post-edit, the system is incrementally updated, by taking into account this sample. Hence, when the system generates the next translation, it will consider the previous user post-edits. It is expected that better translations (or more suited to the human post-editor preferences) will be produced.

In this paper, we evaluate this strategy in an industrial scenario. We study the enhancements brought about by an adaptive system via online learning, and the effects on the post-editing process of data generated by a neural machine translation (NMT) system. To that end, we firstly evaluate our system under laboratory conditions. Next, we conduct the evaluation of the system on a production environment. This experiment involved professional translators, who regularly rely on MT post-editing in their workflow. The results show improvements of adaptive systems in terms of productivity and translation quality. 

\section{Related work}
\label{se:work}

Translation post-editing has been a widely adopted practice in the industry for a long time \citep[e.g.,][]{Vasconcellos85}. As MT technology advanced and improved, the post-editing process gained more relevance and many user studies have demonstrated its capabilities \citep{Aziz12,Bentivogli16,Castilho17,Green13b}.

Adapting an MT system from user post-edits via online learning techniques has also attracted the attention of researchers and industry parallel to the rise of the post-editing protocol. Many advances in this direction were achieved during the CasMaCat \citep{Alabau13} and MateCat \citep{Federico14} projects, which adapted phrase-based statistical machine translation systems incrementally from user post-edits.

Following recent breakthroughs in NMT technology, some works studied the construction of adaptive systems via online learning in this post-editing scenario. \citet{Turchi17} and \citet{Peris17b} proposed to adapt an NMT system with post-edited samples to a new domain via online learning. Other works aimed to refine these adaptation techniques: \citet{Wuebker18} applied sparse updates; \citet{Kothur18} introduced a dictionary of translations for dealing with the novel words included in the new domain. However, in all these works, the users were simulated, due to the economical costs of involving humans within experiments.

User studies on online adaptation from post-edits have been conducted, mainly for phrase-based statistical machine translation systems \citep{Alabau16,Bentivogli16,Denkowski14b,Green13}. Regarding the NMT technology, several user studies have been recently conducted, analyzing different MT technologies \citep{,Koponen19,Jia19} or protocols \citep{Daems19}. The closest work to ours was developed by \citet{Karimova18}, who showed savings in human effort, due to the effect of online learning. But in contrast to our work, the individuals used in \citet{Karimova18} were students, whereas we conducted the study using professional, experienced translators.

\section{Online learning from NMT post-edits}
\label{se:NMT}

NMT relies on the statistical formalization of MT \citep{Brown90}. The goal is to obtain, given a source sentence $\x$, its most likely translation $\hat{\y}$: 

\begin{equation}
\hat{\y} = \argmax_{\y}{\Pr(\y \mid \x)}
\end{equation}

This probability is directly modeled by a neural network with parameters $\VTheta$:

\begin{equation}
\hat{\y} = \argmax_{\y} \log \p(\y \mid \x; \VTheta)
\label{eq:nmt-objective}
\end{equation}

This neural network usually follows an encoder--decoder architecture, featuring recurrent \citep{Bahdanau15,Sutskever14} or convolutional networks \citep{Gehring17} or attention mechanisms \citep{Vaswani17}. The parameters of the model are typically estimated jointly on large parallel corpora, via stochastic gradient descent \citep[SGD;][]{Robbins51,Rumelhart86}. At decoding time, the system obtains the most likely translation by means of a beam search method.

\subsection{Adaption from post-edits via online learning}

During the usage of the MT system, we can leverage the post-edited samples for continuously adapting the system on the fly, as soon as a sentence has been post-edited. This procedure is described in \cref{alg:ol-nmt}: for each sentence to be translated ($\x$), the system produces a translation hypothesis $\hat{\y}$. The user post-edits this sentence, obtaining a corrected version of it ($\y$). Right after this post-editing process, and before translating the next sample, the NMT system is updated, taking into account $\x$ and $\y$.

\begin{algorithm}[h]
    \caption{\label{alg:ol-nmt}Adaptation via online learning during NMT post-editing.}
    \small
    \SetKwInOut{Input}{Input}
    \SetKwInOut{Output}{Output}
    \SetKwInOut{Auxiliar}{Auxiliar}
    \Input{$\VTheta_1$ (initial NMT system), \\
        $\{\x_n\}_{n=1}^{n=N}$ (source sentences) \\
    }
    \Begin{
        $n \leftarrow 1$ \\
        \While{$n \leq N$}{
            $\hat{\y}_n \leftarrow \text{Translate}(\x_n, \VTheta_n)$\\
            $\y_n \leftarrow \text{Post-edit}(\x_n, \hat{\y}_n)$ \\
            $\VTheta_{n+1} \leftarrow \text{Update}((\x_n,\y_n), \VTheta_n)$ \\
            $n \leftarrow n+ 1$ \\
        }
    } 
\end{algorithm} 	

This adaptation of the NMT model can be performed following the same method used in regular training: SGD.

\section{Experimental framework}
\label{se:Exp}
We now describe the experimental conditions arranged in our study: the translation systems and environment, the main features of the tasks under study and the evaluation criteria considered.

\subsection{NMT systems}
Our NMT system was a recurrent encoder--decoder with an additive attention mechanism \citep{Bahdanau15}, built with OpenNMT-py \citep{opennmt}. We used long short-term memory units \citep{Gers00} and we set all model dimensions to $512$. The system was trained using Adam \citep{Adam} with a fixed learning rate of $0.0002$ \citep{Wu16} and a batch size of $60$. We applied label smoothing of $0.1$ \citep{Szegedy15}. At the inference time, we used a beam search with a beam size of 6. We applied joint byte pair encoding to all corpora \citep{Sennrich16}, using $32,000$ merge operations. 

The adaptive systems were built considering the findings from \citet{Peris19}, and conducting an evaluation on a development set. For each new post-edited sample, we performed two plain SGD updates, with a fixed learning rate of $0.05$.
\begin{figure*}[!ht]
    \center
    \includegraphics[width=0.9\textwidth]{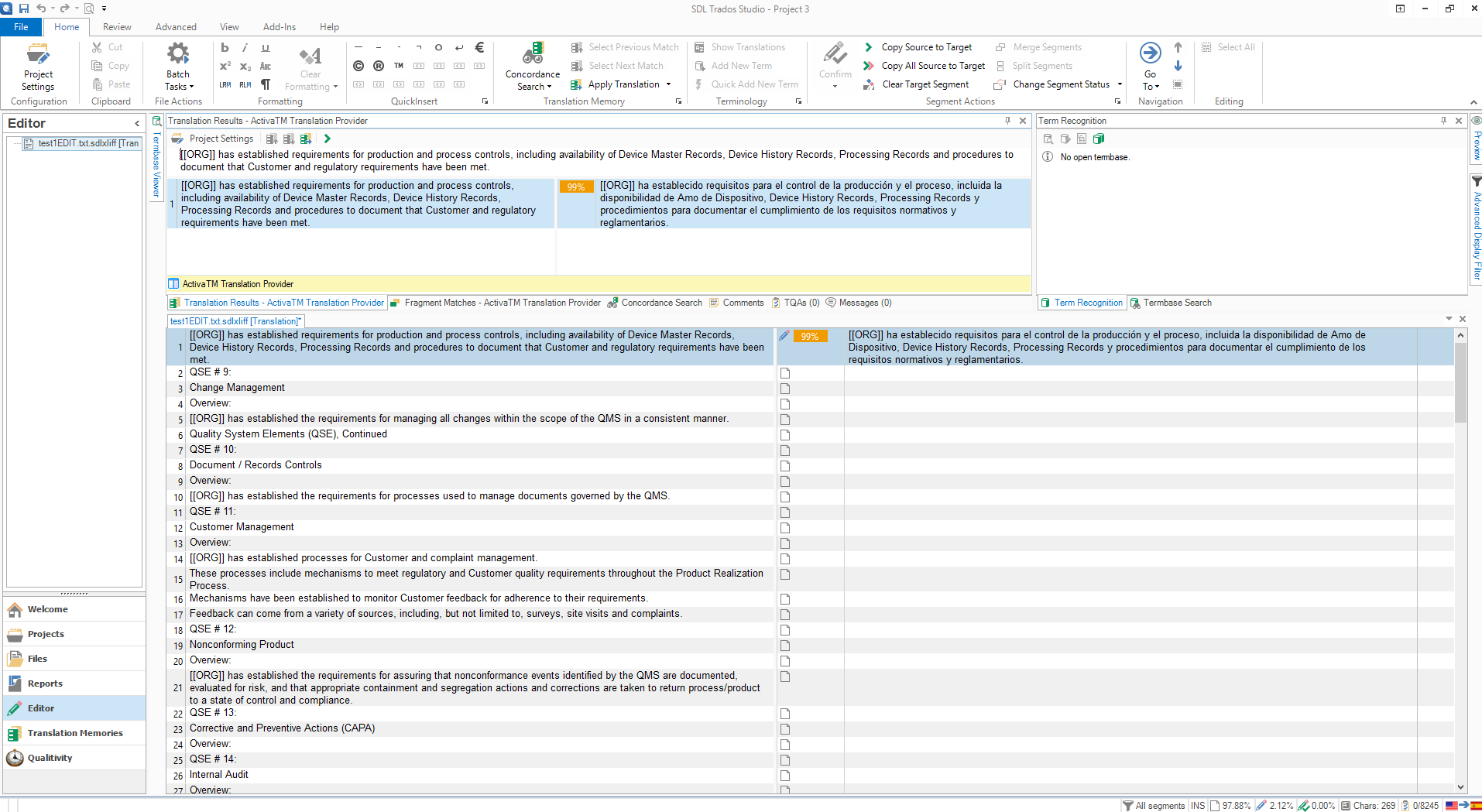}
    \caption{User Interface from SDL Trados Studio. From top to bottom, the first row and the leftmost column correspond to the user menus. On the next row, the middle column contains information about the segment that is being translated: on the left, the source sentence and, on the right, the MT translation. The right column displays the content of the terminological dictionary (if any). The document that is being translated appears on the bottom row: on the left, the original document and, on the right, the user post-edits.
        \label{FigTrados}}
\end{figure*}

\subsection{Translation environment}

In order to assess the benefits of the adaptive system, we started by conducting an experiment with simulated users in a laboratory setting. This study is frequently carried out within the literature \citep[e.g.,][]{Ortiz16}, due to the economical costs of involving humans within experiments. Following common practices, we used the reference sentences as translation post-edits. Therefore, in the static scenario, we assessed the quality of the system using the references. In the adaptive scenario, we translated each source sentence and applied online learning with the corresponding reference.

Once we studied the behavior of the system under simulated conditions, we conducted the experiment with the real users. They were three professional translators, with an average of four years of experience, who regularly make use of MT in their workflow.

The experiment was conducted using SDL Trados Studio 
as the translation environment. This software is widely used in the translation industry, and all the participants use it in their daily work. \cref{FigTrados} shows a screenshot of the SDL Trados Studio interface.

Our NMT system was deployed as a server, which delivered the translations to SDL Trados Studio and performed the adaptation using the post-edits. This system is compatible with all OpenNMT-py models and it is publicly available\footnote{\url{https://github.com/midobal/OpenNMT-py/tree/OnlineLearning}
}. We also developed a plugin that connected {SDL Trados Studio} with our systems.

\subsection{Tasks and evaluation}
\label{SecCorpora}

We evaluated our systems on a real task from our production scenario. This task consisted in a small corpus belonging to a medico-technical domain (description of medical equipments), and was conformed by two documents of 150 sentences each, containing $1.7$ and $2.7$ thousand words respectively. The translation direction was from English to Spanish. Since we lacked an in-domain corpus, we trained a general system with the data from the translation task from WMT'13 \citep{Bojar13}, consisting in $15$ million parallel segments. Next, we applied the FDA data selection technique \citep{Bicici15} for selecting related instances from our general corpus and a medical \citep[UFAL,][]{Bojar17c} and technological\footnote{\url{https://metashare.metanet4u.eu/go2/qtleapcorpus}} ones. We selected $8$ million additional segments, which were used for fine-tuning the general system.

%
%

The effects of adaptivity were assessed according to the post-editing time and to two common MT metrics: (h)BLEU \citep{Papineni02} and (h)TER \citep{Snover06}. For ensuring consistent BLEU scores,  we used sacreBLEU \citep{Post18b}. Since we computed per-sentence BLEU scores, we used exponential BLEU smoothing \citep{Chen14}. In order to determine whether two systems presented statistically significant differences, we applied approximate randomization tests \citep{Riezler05}, with $10,000$ repetitions and a $p$-value of $0.05$.

\section{Results}
\label{se:Res}
As introduced in the previous section, we first analyzed the adaptation process in a simulated environment. Next, we studied and discussed the results obtained in the user trials.

\subsection{Adaptation with simulated users}

\cref{TabSimulations} shows the results in terms of translation quality of a static system, compared with an adaptive one, updated using the reference samples. The results obtained on this synthetic setup support the usefulness of the adaptation via online learning: in all cases, the adaptive system achieved better TER and BLEU than the static one. These differences were statistically significant in all cases but one.  We observed important gains in terms of TER ($5.5$ and $1.1$ points), which suggests a lower human effort required to for post-edit these samples. We also experimented with a larger document ($1,500$ sentences), belonging to the same domain. The adaptation to this larger document was more effective: we observed gains of $10.4$ TER points and $13.6$ BLEU points.

\begin{table}[!ht]
    \centering
    \begin{tabular}{l l l l}
        \toprule
        {\small Test} & {\small System} &  {\small TER [$\downarrow$]} &{\small BLEU [$\uparrow$]}  \\
        \midrule
        \multirow{2}{*}{T1} & Static & $54.0$ & $26.9$ \\
        & Adaptive & $48.5^\dagger$  & $32.0^\dagger$ \\
        \midrule
        \multirow{2}{*}{T2} & Static & $56.1$& $23.4$  \\
        & Adaptive& $55.0$ & $26.3^\dagger$  \\
        \bottomrule
    \end{tabular}
    \caption{\label{TabSimulations}Results of the simulated experiments. Static systems stand for conventional post-editing, without adaptation. Adaptive systems refer to post-editing in an environment with online learning. TER and BLEU were computed against the reference sentences. $^\dagger$ indicates statistically significant differences between the static and the adaptive systems.} 
\end{table}

Additionally to the assessment of the system in terms of translation quality, we need to satisfy an adequate latency, including decoding and updating times. Our NMT system was deployed in a CPU server, equipped with an Intel(R) Xeon(R) CPU E5-2686 v4 at 2.30GHz and 16GB of RAM. On average, generating a translation took the system $0.23$ seconds and each update took $0.45$ seconds. These low latencies allow a correct usage of the system, as the flow of thoughts of the user remains uninterrupted \citep{Nielsen93}.

\subsection{Adaptation with human post-editors}

\begin{table}[h]
    \centering
    \begin{tabular}{l c c}
        \toprule
        User & Static & Adaptive\\
        \midrule
        User 1 & T1 & T2 \\
        User 2 & T2 & T1 \\
        User 3 & T1 & T2 \\
        \bottomrule
    \end{tabular}
    \caption{Distribution of users (1, 2 and 3), test sets (T1 and T2) and scenarios (Static and Adaptive).}
    \label{TabUsersExp}
\end{table}

Once we tested our system in a simulated environment, we moved on to the experimentation with human post-editors. Three professional translators were involved in the experiment. For the adaptive test, all post-editors started the task with the same system, which was adapted to each user using their own post-edits. Therefore, at the end of the online learning process, each post-editor obtained a tailored system. For the static experiment, the initial NMT system remained fixed along the complete process. In order to avoid the influence of translating the same text multiple times, each participant post-edited a different test set under each scenario (static and adaptive), as shown in \cref{TabUsersExp}.

\begin{figure*}[!h]
    \centering
    \begin{subfigure}[b]{0.48\textwidth}
        \includegraphics[width=\textwidth]{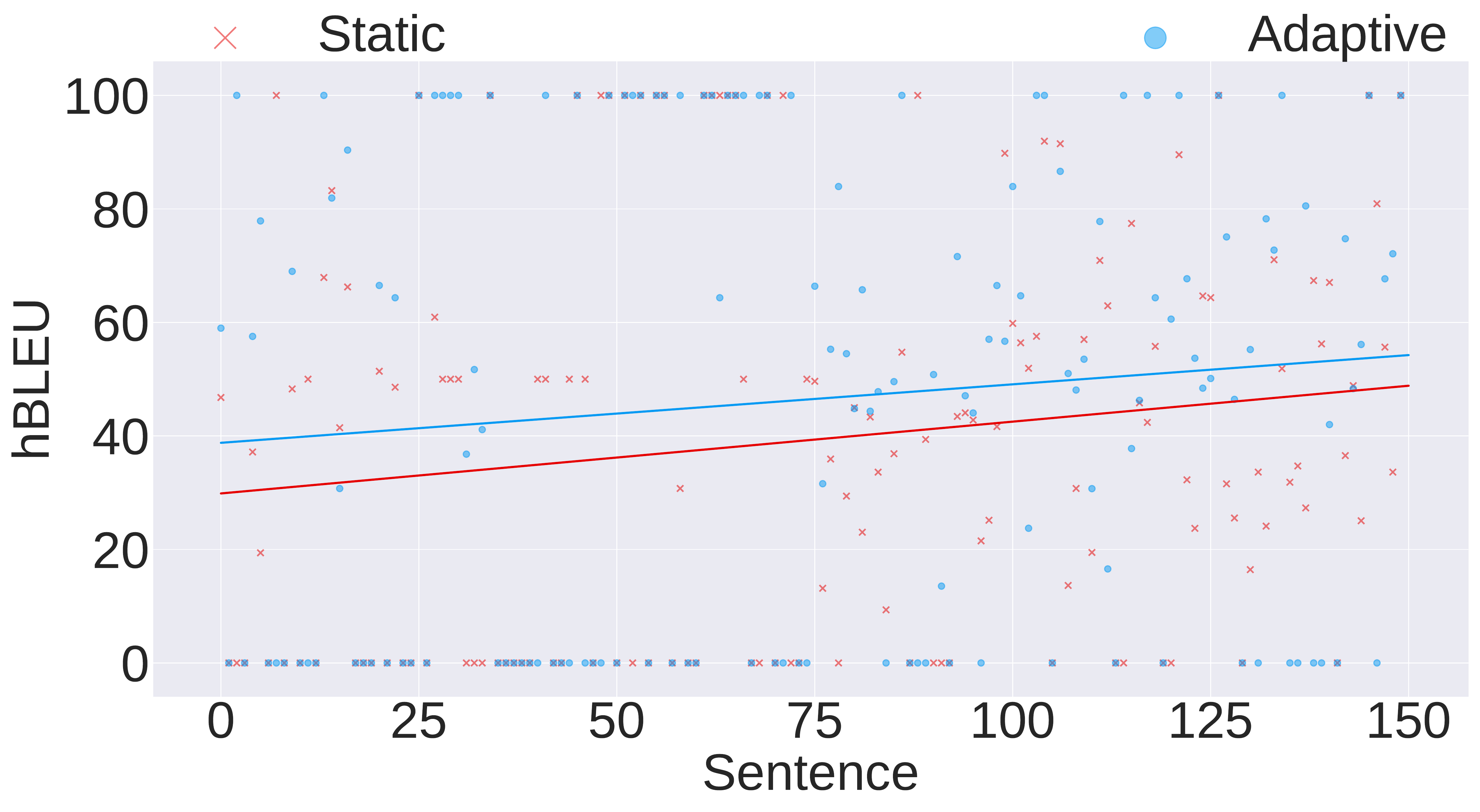}
        \caption{T1.}
        \label{fig:t1-hbleu-per-sentence}
    \end{subfigure}
    \quad  
    \begin{subfigure}[b]{0.48\textwidth}
        \includegraphics[width=\textwidth]{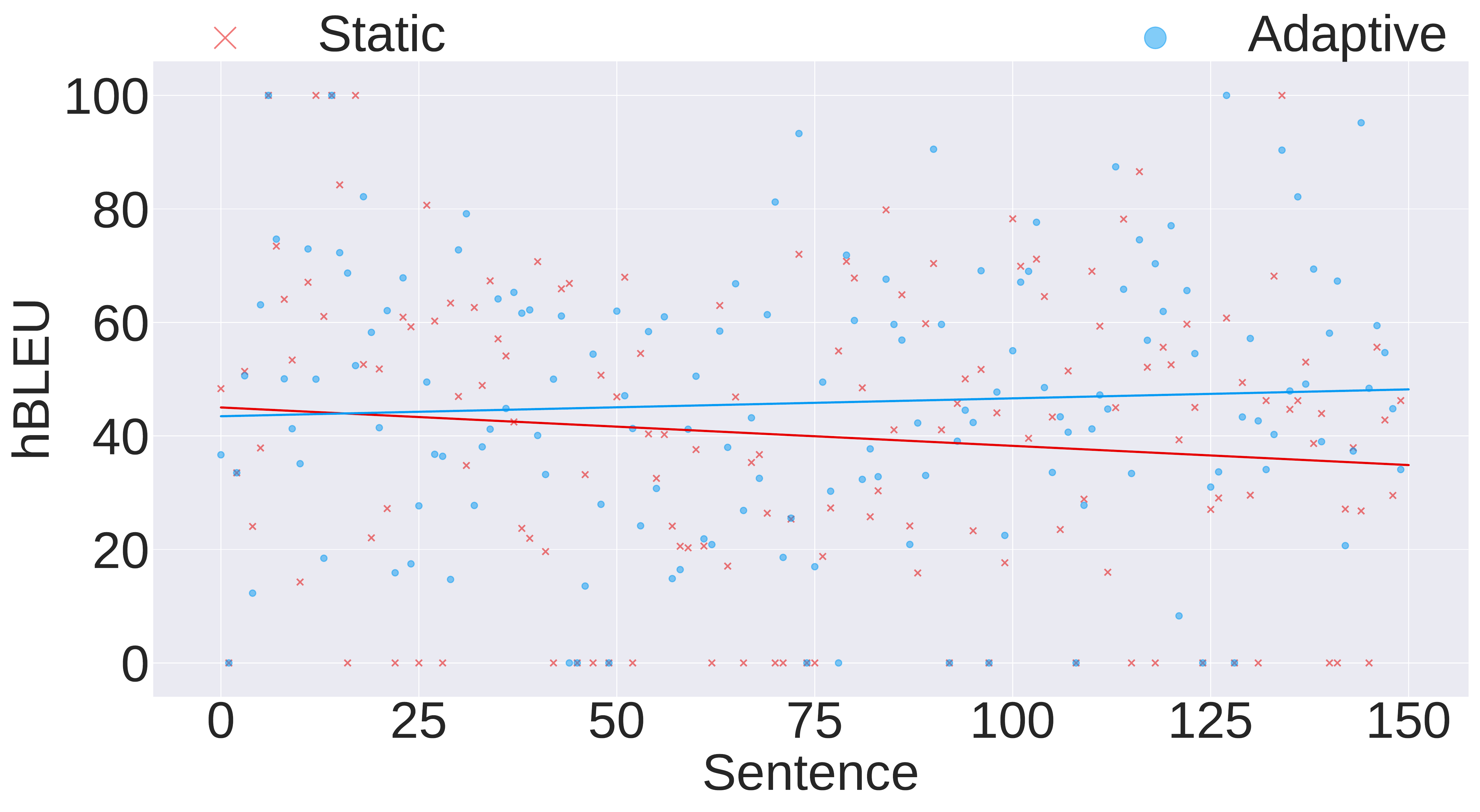}
        \caption{T2.}
        \label{fig:t2-hbleu-per-sentence}
    \end{subfigure}
    \caption[hBLEU per sentence]{\label{fig:hbleu-per-sentence}hBLEU per sentence of static and adaptive systems for both test sets (T1 and T2). Individual sentence scores are plotted for each system, static (red crosses) and adaptive (blue dots). The sentences were processed sequentially, hence, we can observe the progress of the system with its usage. To this end, we show a fit of the scores of each system, in dashed red and solid blue lines, for static and adaptive systems, respectively.}
\end{figure*}

The main results of this experiment are shown in \cref{TabUsers}. These numbers are averages over the results obtained by the different post-editors. 
The large reduction of post-editing time per sentence for the set T1 is especially relevant (an average of $7.5$ seconds per sentence). In the test set T2, the post-editing time of the adaptive system was also slightly lower than the static system one, but only by $0.7$ seconds.

\begin{table}[!ht]
    \centering
    \begin{tabular}{l l c c c}
        \toprule
        {\small Test} & {\small System} & {\small Time (s)}& {\small hTER [$\downarrow$]}  & {\small hBLEU [$\uparrow$]}\\
        \midrule
        \multirow{2}{*}{T1} & Static   & $37.9$  & $39.5$   & $47.3$ \\
        
        & Adaptive & $30.4$  & $34.2$   & $55.1 ^\dagger$\\
        \midrule
        \multirow{2}{*}{T2} & Static   & $45.8$  & $38.4$   & $45.7$ \\
        & Adaptive & $45.1$  & $34.2^\dagger$   & $50.5^\dagger$ \\
        \bottomrule
    \end{tabular}
    \caption{Results of the user experiments. Static systems stand for conventional post-editing, without adaptation. Adaptive systems refer to post-editing in an environment with online learning. Time corresponds to the average post-editing time per sentence, in seconds. hTER and hBLEU refer to the TER and BLEU of the system hypothesis computed against the post-edited sentences. $^\dagger$ indicates statistically significant differences between the static and the adaptive systems.}
    \label{TabUsers}
\end{table}

In terms of translation quality, adaptive systems performed much better than static ones, as reflected by the significant improvements in terms of hTER ($5.3$ and $4.2$ points) and hBLEU ($7.8$ and $4.8$ points). These results show that adaptive systems generated more correct translations, as they required less post-edits from the user. 

In order to gain additional insights into the adaptation process, we studied the evolution of the hBLEU during the post-editing  process. To this end, \cref{fig:hbleu-per-sentence} compares the hBLEU per sentence of static and adaptive systems, for both test sets. Since the sentences were processed sequentially, we study the progress of the systems along its usage: for observing these trends, we computed a linear fit of the scores of each system via the least squares method.

In \cref{fig:t1-hbleu-per-sentence}, we observe that for the test split T1, the adaptive system consistently produced slightly better hypotheses than the static one, but there was no clear evidence on the effects of online learning. Both systems behaved similarly: the hBLEU values were gradually increased, which suggests either that the test document was increasingly easy to translate or that the user felt more comfortable with the style and translations provided by the system. Therefore, they applied less post-edits to the final sentences.

In the case of T2 (\cref{fig:t2-hbleu-per-sentence}), we observe a degradation on the hBLEU of the static system, as the post-editing process advances. This degradation is prevented by the adaptive system, in which the hBLEU is even slightly increased. The effects of the adaptation are noticeable from the $30$th sentence onwards.

Finally, it is interesting to compare the simulated experiment against this one. We observed that, in terms of automatic metrics, the system yielded much better results when evaluating against post-edits, rather than against reference sentences (compare the ``Static'' rows from \cref{TabUsers} and \cref{TabSimulations}, respectively). This suggests that the translation hypotheses provided by the system were useful to the human users, as they produced similar post-edited samples. It is also worth to point out that the adaptation process was, in most cases, slightly less effective in the simulated experiment.

\subsection{User perceptions and opinions}
After finishing each experiment, the participants answered a questionnaire regarding the post-editing task they had just performed. In this survey, we asked the users about their level of satisfaction of the translations they produced, whether they preferred to perform post-editing or translating from scratch and their opinions on the automatic translations provided, in terms of grammar, style and overall quality. We also requested for them to give their feedback on the task, as an open-answer question. 

The users were generally satisfied with the translations they generated. In all cases, they preferred to perform this translation task via post-editing rather than translating from scratch. Two of them preferred to perform this translation from scratch in less than a $25\%$ of the sentences. The other post-editor preferred to translate from scratch around a $50\%$ of the sentences. In all cases, they are keen to perform translation post-editing in the future. These perceptions on the MT utility are slightly better than those reported by \citet{Daems19}. We believe that these differences are due to the background in translation post-editing that our users had: they perform translation post-editing as their regular way of work; therefore, they perceptions toward this methodology are generally favorable.

Regarding the translation quality offered by the NMT system, their general opinion is that the system produced translations of average quality. The strongest attribute of the translations was their grammatical accuracy. The style and overall quality was perceived in some cases below the average, depending on the user and the experimental condition.

In order to avoid biases, the users did not know whether the experiment they performed featured a static or an adaptive system. Once they finished both experiments, they were asked to identify the adaptive systems. All users guessed correctly which one was the adaptive system.

Regarding their general opinions, they all observed how corrections applied on one segment were generally reflected in the following segment, especially corrections related to product names, grammatical structures and lexical aspects. This mostly reduced upcoming corrections to changes in the style. Overall, their perception was that the static system produced less fluent translations, and that the machine translation was very good in most cases, but useless in a few ones.

The post-editors reported a couple of minor issues regarding the NMT system: in a few cases, they noticed that a domain-specific term was ``forgotten'' by the system, being wrongly translated. In addition, the users noticed in some cases, the occurrence of some made-up words (e.g., ``absolvido''). This problem was probably caused by an incorrect segmentation of a word, via the byte pair encoding process. In order to deploy natural and effective translation systems, these problems need to be addressed.

\section{Conclusions and future work}
\label{se:Con}

We conducted an evaluation of an adaptive NMT system in a post-editing scenario. The system leveraged the data generated during the post-editing process for adapting its underlying models. After testing the system in a laboratory setting, we conducted an experiment involving three professional translators, who regularly make use of MT post-editing. We observed reductions in post-editing times and significant improvements in terms of hTER and hBLEU, due to online learning. The users were pleased with the system. They noticed that corrections applied on a given segment generally were reflected on the successive ones, making the post-editing process more effective and less tedious.

As future work, we should address some of the concerns noticed by the post-editors, namely, the degradation of domain-specific terms and the incorrect generation of words due to subwords. To that end, we should study and analyze the hypotheses produced by the adaptive system and the post-edits performed by the users, similarly as \citet{Koponen19}. Moreover, we want to integrate our adaptive systems together with other translation tools, such as translation memories or terminological dictionaries, with the aim of fostering the productivity of the post-editing process. With this feature-rich system, we would like to conduct additional experiments involving more diverse languages and domains, using domain-specialized NMT systems, testing other models \citep[e.g., Transformer,][]{Vaswani17} and involving a larger number of professional post-editors. Finally, we also intend to implement the interactive--predictive machine translation protocol \citep{Lam18,Peris19} in our translation environment, and compare it with the regular post-editing process.

\section*{Acknowledgements}
The research leading to these results has received funding from the Spanish Centre for Technological and Industrial Development (Centro para el Desarrollo Tecnol{\'o}gico Industrial) (CDTI) and the European Union through Programa Operativo de Crecimiento Inteligente (Project IDI-20170964). We gratefully acknowledge the support of NVIDIA Corporation with the donation of a GPU used for part of this research, and the translators and project managers from Pangeanic for their help with the user study.

\bibliographystyle{apalike}
\bibliography{OL}

\end{document}